\documentclass[sigconf]{acmart}

\def\BibTeX{{\rm B\kern-.05em{\sc i\kern-.025em b}\kern-.08emT\kern-.1667em\lower.7ex\hbox{E}\kern-.125emX}}
    
\begin{document}

\acmSubmissionID{328}

\def\method{Div\-Graph\-Pointer} 
\def\nodivmethod{Graph\-Pointer}
\def\rnnmethod{Seq\-Pointer} 

%
\title[DivGraphPointer]{DivGraphPointer: A Graph Pointer Network for Extracting Diverse Keyphrases}

\copyrightyear{2019} 
\acmYear{2019} 
\setcopyright{acmlicensed}
\acmConference[SIGIR '19]{Proceedings of the 42nd International ACM SIGIR Conference on Research and Development in Information Retrieval}{July 21--25, 2019}{Paris, France}
\acmBooktitle{Proceedings of the 42nd International ACM SIGIR Conference on Research and Development in Information Retrieval (SIGIR '19), July 21--25, 2019, Paris, France}
\acmPrice{15.00}
\acmDOI{10.1145/3331184.3331219}
\acmISBN{978-1-4503-6172-9/19/07}

%

\author{Zhiqing Sun}
\affiliation{
    \institution{Peking University}
    \city{Beijing}
    \country{China}
}
\email{1500012783@pku.edu.cn}

\author{Jian Tang}
\affiliation{
    \institution{Mila-Quebec Institute for Learning Algorithms}
    \institution{HEC Montr\'eal}
    \institution{CIFAR AI Research Chair}
    \city{Montr\'eal}
    \country{Canada}
}
\email{jian.tang@hec.ca}

\author{Pan Du}
\affiliation{
    \institution{Universit\'e de Montr\'eal}
    \city{Montr\'eal}
    \country{Canada}
}
\email{pandu@iro.umontreal.ca}

\author{Zhi-Hong Deng}
\affiliation{
    \institution{Peking University}
    \city{Beijing}
    \country{China}
}
\email{zhdeng@pku.edu.cn}

\author{Jian-Yun Nie}
\affiliation{
    \institution{Universit\'e de Montr\'eal}
    \city{Montr\'eal}
    \country{Canada}
}
\email{nie@iro.umontreal.ca}



 




%

%
\begin{abstract}
Keyphrase extraction from documents is useful to a variety of applications such as information retrieval and document summarization. This paper presents an end-to-end method called DivGraphPointer for extracting a set of diversified keyphrases from a document.  DivGraphPointer combines the advantages of traditional graph-based ranking methods and recent neural network-based approaches. Specifically, given a document, a word graph is constructed from the document based on word proximity and is encoded with graph convolutional networks, which effectively capture document-level word salience by modeling long-range dependency between words in the document and aggregating multiple appearances of identical words into one node. Furthermore, we propose a diversified point network to generate a set of diverse keyphrases out of the word graph in the decoding process. Experimental results on five benchmark data sets show that our proposed method significantly outperforms the existing state-of-the-art approaches. 
\end{abstract}

%
%


%

\maketitle

\section{Introduction}

Keyphrase extraction from documents is  useful in a variety of tasks such as information retrieval \cite{kim2013applying}, text summarization \cite{qazvinian2010citation}, and question answering \cite{li2010question}. It allows to identify the salient contents from a document. The topic has attracted a large amount of work in the literature.

Most traditional approaches to keyphrase extraction are unsupervised approaches. They usually first identify the candidate keyphrases with some heuristics (e.g., regular expressions), and then rank the candidate keyphrases according to their importance in the documents \cite{hasan2014automatic}. Along this direction, the state-of-the-art algorithms are graph-based ranking methods \cite{mihalcea2004textrank,wan2008single,liu2010automatic}, which first construct a word graph from a document and then determine the importance of the keyphrases with random walk based approaches such as PageRank \cite{brin1998anatomy}. By constructing the word graph, these methods can effectively identify the most salient keyphrases. Some diversification mechanisms have also been investigated in some early work \cite{mei2010divrank,bougouin2013topicrank} to address the  problem of over-generation of the same concepts in keyphrase extraction. However, these methods are fully unsupervised. They rely heavily on manually designed heuristics, which may not work well when applied to a different type of document.
In  experiments, we also observe that the performance of these methods is usually limited and inferior to the supervised ones.

Recently, end-to-end neural approaches for keyphrase extraction have been attracting growing interests \cite{zhang2016keyphrase,meng2017deep,zhang2017deep}. The neural approaches usually studied keyphrase extraction in the encoder-decoder framework \cite{sutskever2014sequence}, which first encodes the input documents into vector representations and then generates the keyphrases with Recurrent Neural Networks (RNN) \cite{mikolov2010recurrent} or CopyRNN \cite{gu2016incorporating} decoders conditioned on the document representations. These neural methods have achieved state-of-the-art performance on multiple benchmark data sets with end-to-end supervised training. The end-to-end training offers a great advantage that the extraction process can adapt to the type of documents. However, compared to the unsupervised graph-based ranking approaches, existing end-to-end approaches only treat documents as sequences of words. They do not benefit from the a more global graph structure that provides useful
document-level word salience information such as  long-range dependencies between words, as well as a synthetic view on the multiple appearances of identical words in the document. Another problem of these end-to-end methods is that they cannot guarantee the diversity of the extracted key phrases: it is often the case that several similar keyphrases are extracted. Therefore, we are seeking an approach that can have the advantage of modeling document-level word salience, generating diverse keyphrases, and meanwhile  be efficiently trained in an end-to-end fashion. 

\begin{figure*}[t]
    \centering
    \includegraphics[width=0.95\textwidth]{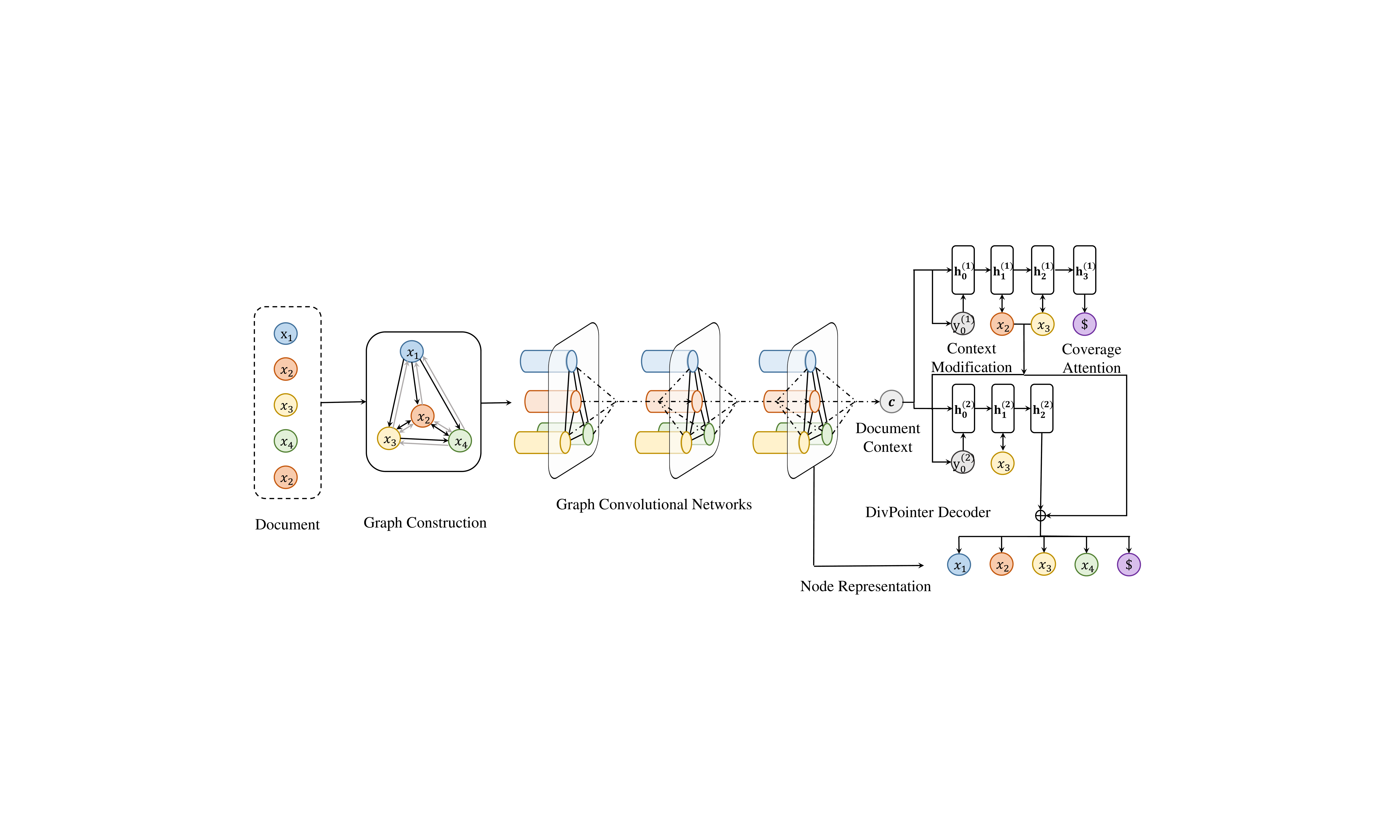}
    \caption{Illustration of our encoder-decoder architecture for keyphrase extraction. In this example, the document is a sequence of words, namely, $d = \langle x_1,x_2,x_3,x_4,x_2 \rangle$, and we have generated the first keyphrase $\textit{y}^{1} = \langle x_2,x_3 \rangle$. We are predicting $\textit{y}^2_2$, the second word for keyphrase $\textit{y}^2$, which will be selected from the nodes within the graph and the ending token $\$$  for a keyphrase. Note that multiple appearances of $x_2$ in the document is aggregated into only one node in the constructed word graph. (Better viewed in color)}
    \label{fig:structure}
\end{figure*}

In this paper, we propose an end-to-end approach called \method{} for extracting diversified keyphrases from documents. Specifically, given an input document, we first construct a word graph from it, which aggregates identical words into one node and captures both the short- and long-range dependency between the words in the document. Afterwards, the graph convolutional neural network \cite{kipf2016semi} is applied to the word graph to learn the representations of each node, which effectively models the word salience. To extract diverse keyphrases from documents, we propose a diversified pointer network model \cite{vinyals2015pointer} over the word graph, which dynamically picks nodes from the word graph to construct the keyphrases. Two diversity mechanisms are proposed to increase the diversity among the generated keyphrases.
Specifically, we employ a coverage attention mechanism \cite{tu2016modeling} to address the over-generation problem in keyphrase extraction at lexical level and a semantic modification mechanism to dynamically modify the encoded document representation at semantic level. Figure \ref{fig:structure} illustrates our approach schematically.
The whole framework can be effectively and efficiently trained with back-propagation in an end-to-end fashion. Experimental results show that our proposed \method{} achieves state-of-the-art performance for keyphrase extraction on five benchmarks and significantly outperforms the existing supervised and unsupervised keyphrase extraction methods.

The contribution of this paper is twofold:
\begin{itemize}
    \item We propose a graph convolutional network encoder for keyphrase extraction that can effectively capture document-level word salience.
    \item We propose two complementary diversification mechanisms that help the pointer network decoder to extract diverse keyphrases.
\end{itemize}
\section{Related Work}
The most traditional keyphrase extraction method is based on Tf-Idf \cite{sparck1972statistical}: It identifies words that appear frequently in a document, but do not occur frequently in the entire document collection. These words are expected to be salient words for the document. The same idea can be applied on a set of keyphrase candidates identified by some syntactic patterns \cite{frank1999domain,hulth2003improved}. 
However, the drawback of this family of approaches is that each word (or phrase) is considered in isolation. The inherent relations between words and between phrases are not taken into account. Such relations are important in keyphrase extraction.

To solve this problem, approaches based on word graphs have been proposed. In a word graph, words are connected according to some estimated relations such as co-occurrences. A graph-based extraction algorithm can then take into account the connections between words.  The TextRank algorithm \cite{mihalcea2004textrank} was the first graph-based approach for keyphrase extraction. Given a word graph built on co-occurrences,
it calculates the importance of candidate words with PageRank. The importance of a candidate keyphrase is then estimated as the sum of the scores of the constituent words. 
Following this work, the DivRank algorithm \cite{mei2010divrank} was proposed to balance the importance and diversity of the extracted keyphrases.
%
The TopicRank algorithm \cite{bougouin2013topicrank} was further proposed for topic-based keyphrase extraction. This algorithm first clusters the candidate phrases by topic and then chooses one phrase from each topic, which is able to generate a diversified set of keyphrases. In TopicRank, a complete topic graph is constructed to better capture the semantic relations between topics.
The graph-based document representation can effectively model document-level word salience. However, these methods are fully unsupervised: the way that keyphrases are identified from a word graph is designed manually. Such a method lacks the flexibility to cope with different types of documents. In our proposed methods, we will use an end-to-end  supervised training in order to adapt the extraction process to documents.
In experiments, this also yields better performance.


Indeed, end-to-end neural approaches to keyphrase extraction have attracted a growing attention in recent studies. \citet{zhang2016keyphrase} treated keyphrase extraction as a sequence labeling task and proposed a model called joint-layer RNN for extracting keyphrases from Twitter data. \citet{meng2017deep} first proposed an encoder-decoder framework for keyphrase extraction. However, their RNN-based encoder and decoder treat a document as a sequence and ignore the correlation between keyphrases. Afterwards, \citet{zhang2017deep} further proposed a CNN-based model for this task. The copy mechanism \cite{gu2016incorporating} is employed to handle the rare word problem in these encoder-decoder approaches, which allows to copy some words from the input documents. All the above approaches are based on word sequences, which inherit the well known problem that only local relations between words can be coped with. Compared to them, our model encodes documents with graphs, which are able to model more global document-level word salience. 






Deep graph-based methods have been used for other tasks. They are also relevant to our work. For example, \citet{yasunaga2017graph} proposed to use graph convolutional networks to rank the salience of candidate sentences for document summarization. \citet{marcheggiani2017encoding} studied encoding sentences with graph convolutional neural networks for semantic role labeling. \citet{bastings2017graph} studied graph convolutional encoders for machine translation. In this paper, we target a different task. This is the first time that graph convolutional neural networks are used for keyphrase extraction. 

Diversity is an important criterion in keyphrase extraction: it is useless to extract a set of similar keyphrases. Diversity has been studied in IR for search result diversification \cite{carbonell1998use,santos2010exploiting}. 
Two main ideas have been used: selecting results that are different from those already selected; or selecting a set of results that ensure a good coverage of topics.
In keyphrase extraction, there are also a few attempts to deal with diversity. 
Similar to our work, \citet{zhang2018keyphrase} and \citet{chen2018keyphrase} also employed a coverage mechanism to address the diversification problem. \citet{chen2018keyphrase} further proposed a review mechanism to explicitly model the correlation between the keyphrases. However, their approaches are different from ours in multiple ways. 
First, this paper focuses on keyphrase extraction, while their approaches focus on keyphrase generation, which also requires generating keyphrases that cannot be found in the document. 
Furthermore, we use a hard form of coverage attention mechanism that penalizes the attention weight by one-hot vectors, while they used the same form in machine translation as the original paper \cite{tu2016modeling}. We believe that in the keyphrase extraction task where source and target share the same vocabulary, a hard coverage attention could avoid the error propagation of attentions. Moreover, by directly applying the coverage attention on the word graph, we efficiently penalize all appearances of identical words.
Finally, the review mechanism was employed in the decoding phase of RNN in \citet{chen2018keyphrase}, while our context modification mechanism modifies the initial semantic state of RNN. We expect that our context modification and coverage attention mechanisms can explicitly address the over-generation problem in keyphrase extraction at both semantic level and lexical level, and thus are complementary to each other. In contrast, the review mechanism and the coverage mechanism used in \citet{chen2018keyphrase} provide both an attentive context, which tend to play a similar role and are somehow duplicated.
\section{Deep Graph-based Diversified Keyphrase Extraction}
Our Diversified Graph Pointer (\method{}) is based on the encoder-decoder framework. We first construct a graph from a document from word adjacency matrices and encode the graph with the Graph Convolutional Networks (GCN) \cite{kipf2016semi}. Afterwards, based on the representation of the graph, we generate the keyphrases one by one with a decoder. Note that during the generation process, we restrict the output to the nodes of the graph. In other words, we only extract words appearing in the documents. Figure \ref{fig:structure} presents an illustration of the \method{} framework. 

\subsection{Graph Convolutional Network Encoder}

In this part, we present our graph-based encoders.  Traditional unsupervised graph-based ranking approaches for keyphrase extraction have shown promising performance at estimating the salience of words, which motivated us to develop deep graph-based encoders. Compared to sequence-based encoders such as RNN and CNN, graph-based encoders have several advantages. For example, graph-based encoders can explicitly leverage the short- and long-term dependency between words. Moreover, while the sequence-based encoders treat different appearances of an identical word independently, the graph representation of document naturally represents the identical word at different positions as a single node in the word graph. In this way, the graph-based encoding approaches can aggregate the information of all appearances of an identical word when estimating its salience, and thus employ a similar idea as in Tf-Idf  \cite{sparck1972statistical} and PageRank \cite{brin1998anatomy} methods: important words tend to be more frequent in the document and widely linked with other important words.

\subsubsection{Graph Construction}
Our model represents a document by a complete graph in which identical words are nodes and edges are
weighted according to the strength of the structural relations between nodes.
Instead of manually designing and extracting connections between words (e.g. based on co-occurrence weights), we rely on the basic proximity information between words in the sentences, and let training process learn to use it. The basic assumption is that the closer two words in a sentence, the stronger their relation. 
Specifically, we construct a directed word graph from the word sequences of a document in both directions. The adjacency matrices are denoted as: $\overleftarrow{A}$ and $\overrightarrow{A}$.
We assume that two words are related if they appear close to each other in a document. This extends the traditional adjacency relation to a more flexible proximity relation. The strength of the relation between a pair of words depends on their distance. 
Specifically, similar to \cite{bougouin2013topicrank}, we define the weight from word $w_i$ to word $w_j$ in the two graphs as follows:    
\begin{align}
\overleftarrow{A}_{ij} &= \sum_{p_i \in \mathcal{P}(w_i)}  \sum_{p_j \in \mathcal{P}(w_j)} relu \left(\frac{1}{p_i - p_j} \right)\\
\overrightarrow{A}_{ij} &= \sum_{p_i \in \mathcal{P}(w_i)}  \sum_{p_j \in \mathcal{P}(w_j)} relu \left(\frac{1}{p_j - p_i} \right),
\end{align}
where $\mathcal{P}(w_i)$ is the set of the position offset $p_i$ of word $w_i$ in the document. The function $relu(\cdot) = max(\cdot, 0)$  aims to filter the uni-directional information and help the graph-based encoder focusing on the order of the sequence.

In order to stabilize the iterative message propagation process in graph convolutional network encoder, we normalize each adjacency matrix $A\in\{\overleftarrow{A}, \overrightarrow{A}\}$ by $\hat{A} = \tilde{D}^{-\frac{1}{2}} \tilde{A} \tilde{D}^{-\frac{1}{2}}$, 
where $\tilde{A} = A + I_N$ is the adjacency matrix $A$ with self-connections, and $ \tilde{D} = \sum_{j} \tilde{A}_{ij}$ is the degree matrix. The purpose of this re-normalization trick \cite{kipf2016semi} is to constrain the eigenvalues of the normalized adjacency matrices $\overleftarrow{\hat{A}}, \overrightarrow{\hat{A}}$ close to 1.

Such a graph construction approach differs from the one used in TextRank \cite{mihalcea2004textrank} that captures the co-occurrence in a limited window of words. $\overleftarrow{\hat{A}} + \overrightarrow{\hat{A}}$ forms a complete graph where all nodes are interconnected. As stated in \cite{bougouin2013topicrank}, the completeness of the graph has the benefit of providing a more exhaustive view of the relations between words. Also, computing weights based on the distances between offset positions bypasses the need for a manually defined parameter such as window size.

\subsubsection{Graph Convolutional Networks}
Next, we encode the nodes with multi-layer Graph Convolutional Networks (GCNs) \cite{kipf2016semi}. Each graph convolutional layer generally consists of two stages. In the first stage, each node aggregates the information from its neighbors; in the second stage, the representation of each node is updated according to its current representation and the information aggregated from its neighbors. 
Given the node representation matrix $H_{l}$ in the $l$-th layer, the information aggregated from the neighbors, denoted as $f_l\left(H_{l}\right)$, can be calculated as follows: 
\begin{equation}
f_l\left(H_{l}\right) = \overleftarrow{\hat{A}} H_l\overleftarrow{W}_l +
\overrightarrow{\hat{A}} H_l \overrightarrow{W}_l + H_lW_l
\end{equation}
which aggregates the information from both the neighbors defined in the two matrices and the node itself. Here, $\overleftarrow{W}_l, \overrightarrow{W}_l, W_l$ are layer-specific trainable weight matrices.

Once the information from the neighbors are aggregated, inspired by \citet{he2016deep} and \citet{gehring2017convolutional}, we updated the node representation with a residual Gated Linear Unit (GLU) activation:
\begin{equation}
H_{l+1} = H_l + f_l\left(H_l\right) \otimes \sigma\left(g_l\left(H_l\right)\right)
\end{equation}
where $\sigma$ is the sigmoid function and $\otimes$ is the point-wise multiplication. $g_l$ is another function defined in a similar way as $f_l$, which is used as the gating function of the information collected from the neighbors. $H_0$ is initialized with the pretrained word embedding matrix, and the residual connection is omitted in its activation.


The representation of the entire graph (or the document representation) $\mathbf{c}$ is then obtained by averaging the aggregation of the last layer's node representations $f_L(H_L)$, where $L$ denotes the total number of GCN layers.

Based on the encoded document representation $\mathbf{c}$, we propose a decoder, named DivPointer, to generate summative and diverse keyphrases in the next section.

\subsection{Diversified Pointer Network Decoder}
In this part, we introduce our approach of keyphrase extraction based on the graph representation. Most previous end-to-end neural approaches select keyphrases independently during the decoding process. However, ignoring the diversity among phrases may lead to multiple similar keyphrases, undermining the representativeness of the keyphrase set. Therefore, we propose a DivPointer Network with two mechanisms on semantic level and lexical level respectively to improve the diversity among keyphrases during the decoding process.

\subsubsection{Pointer Network}

The decoder is used to generate output keyphrases according to the representation of the input document. We adopt a pointer network \cite{vinyals2015pointer}  with diversity enabled attentions to generate keyphrases. A pointer network is a neural architecture to learn the conditional probability of an output sequence with elements that are discrete tokens corresponding to positions in the original data space. The graph nodes corresponding to words in a document are regarded as the original data space of the pointer network in our case.



Specifically, the pointer decoder receives the document representation $\mathbf{c}$ as the initial state $\mathbf{h}_0^{(i)}$, and predicts each word $\textit{y}_t^{(i)}$ of a keyphrase $\textit{y}^{(i)}$ sequentially based on $\mathbf{h}_t^{(i)}$:
\begin{equation}
\mathbf{h}_t^{(i)} = RNN(\mathbf{y}_{t-1}^{(i)}, \mathbf{h}_{t-1}^{(i)})
\end{equation}
where $\mathbf{y}_{t-1}^{(i)}$ denotes the node representation of the word $\textit{y}_{t-1}^{(i)}$ that keyphrase $\textit{y}^{(i)}$ generated at the previous step. $\mathbf{h}_t$ is the hidden state of an RNN. The word $\textit{y}_t^{(i)}$ is then selected with a pointer network according to certain attention mechanism based on $\mathbf{h}_t^{(i)}$. 

A general attention \cite{bahdanau2014neural} score $e_{t,j}$ on each graph node $\textit{x}_{j\leq N}$ with respect to the hidden state $\mathbf{h}_t^{(i)}$ can be computed by:
\begin{equation}
e_{t,j}^{(i)} = \mathbf{v}^T \tanh (W_h \mathbf{h}_t^{(i)} + W_x \mathbf{x}_j + \mathbf{b})
\end{equation}
where $\mathbf{x}_j$ is the node representation of $\textit{x}_j$ taken from $H_L$ and $\mathbf{v}^T$, $W_h$, $W_x$, and $\mathbf{b}$ are parameters to be learned.
We can then obtain the pointer distribution on the nodes by normalizing $\{e_{t,j}^{(i)} \}$:
\begin{equation}
p(\textit{y}_t^{(i)} = \textit{x}_j) = \frac{\exp (e_{t,j}^{(i)})}{\sum_{k = 1}^N \exp (e_{t,k}^{(i)})}
\end{equation}
With this distribution, we can select a word with the maximum pointer probability as $\textit{y}_t^{(i)}$ from the graph nodes. 

However, the attention mechanism above is merely built on the global hidden state, namely the document representation $\mathbf{c}$. Diversity among the generated keyphrases can hardly be addressed without taking the previously generated keyphrases into consideration in the decoding process. Two mechanisms aiming at achieving diversity among keyphrases generated are introduced in the following sections.




\subsubsection{Context Modification}
During the decoding process, the document representation $\mathbf{c}$ is the key to maintain a global semantic context for each word to be generated. However, the constant context may lead to generating similar keyphrases repeatedly, which hurts the representativeness of the generated keyphrase set in reality. 

Therefore, we propose to update the context $\mathbf{h}_0^{(i)}$ dynamically based on the previously generated keyphrases $\textit{y}^{(1:i-1)}$ while decoding the words for the $i^{th}$ phrase $\textit{y}^{(i)}$, as follows:


\begin{align}
& \mathbf{y}_0^{(i)} = W_y \left[\mathbf{c}, \overline{\mathbf{y}^{(1:i-1)}}\right] + \mathbf{b}_y\\
& \mathbf{h}_0^{(i)} = \tanh \left(W_h \left[\mathbf{c}, \overline{\mathbf{y}^{(1:i-1)}}\right] + \mathbf{b}_s\right)
\end{align}
where we introduce the average representation of the previous generated keyphrases  $\overline{\mathbf{y}^{(1:i-1)}}$ into the model to learn a updated context. $\overline{\mathbf{y^{(:)}}}$ is initialized as $\mathbf{0}$. $\mathbf{y}_0^{(0)}$  and $\mathbf{h}_0^{(0)}$ are initialized accordingly.

We find the modified context helps the pointer network to focus on the keyphrases with different meanings yet to come. Therefore, it can solve the over-generation problem in the previous deep generation models.

\subsubsection{Coverage Attention}
In addition to the context modification mechanism, we also adopt the coverage mechanism to enhance diversity among keyphrases on lexical level. Coverage mechanism has been well investigated in machine translation \cite{tu2016modeling}, search result diversification \cite{santos2010exploiting} and document summarization \cite{see2017get}. When the criterion is used in a neural network, it generally maintains a coverage vector to keep track of the attention history. More specifically, they directly use the sum of previous alignment probabilities as the coverage for each word in the input.

Since keyphrases are usually short and summative, their meanings are more sensitive to the term replacement than those of sentences or documents. Based on this observation, we propose a coverage attention on lexical level with the help of one-hot representations of the previously generated keyphrases. Specifically, we use the sum of the one-hot vectors of the previously generated keyphrases as a coverage representation.
\begin{equation}
c_j^{(i)} = c_j^{(i-1)} + \sum_{t} o_{t,j}^{(i-1)}
\end{equation}
where $c_j^{(i)}$ is the coverage value of the $j^{th}$ node for the $i^{th}$ keyphrase and $o_{t,j}^{(i)}$ is the $j^{th}$ element of the $t^{th}$  one-hot vector in the $i^{th}$ keyphrase. It is $1$ if the $t^{th}$ generated word is the $j^{th}$  node in the graph, $0$ otherwise.

The coverage representation is then incorporated into the attention mechanism of our DivPointer network as follows:
\begin{equation*}
e_{t,j}^{(i)} = \mathbf{v}^T \tanh (W_h \mathbf{h_t}^{(i)} + W_x \mathbf{x_j} + \mathbf{w}_c c_j^{(i)} + \mathbf{b})
\end{equation*}

Note that our coverage attention mechanism differs from the original form \cite{tu2016modeling}  that was designed for machine translation. The reason for the change is that in the keyphrase extraction task, the source and the target share the same vocabulary. So a hard coverage attention could avoid the error propagation of attentions to similar words. Instead, the soft attention distribution cannot precisely represent the previous generated keyphrases.

\section{Training and Decoding}
In both training and decoding, the $\$$ symbol is added to the end of the keyphrases to help learn to stop adding more words. The whole model is trained to maximize the log-likelihood of the words in the keyphrases according to the given input document. Specifically, the training objective is defined as below:
\begin{equation}
\begin{split}
p(\textit{y}^{(i)}) &= \prod p (\textit{y}_{j}^{(i)} | \textit{y}_{1:j-1}^{(i)}, \textit{y}^{(1:i-1)}, \mathbf{c})\\
L  &= - \frac{1}{\sum K_d} \sum_{d=1}^{D} \sum_{k = 1}^{K_d} \log p (\textit{y}^{(d, k)})
\end{split}
\end{equation}
where  $D$  is the number of document and $K_d$ is the number of keyphrases in each document. $p(\textit{y}^{(i)})$ is the generative probability of the $i^{th}$ keyphrase given the previous generated keyphrases and the context $\mathbf{c}$. For simplicity, we omit the conditional notation here. $\textit{y}^{(d, k)}$ is the $k^{th}$ keyphrase in the $d^{th}$ document, which is also conditioned on $\textit{y}^{(d, 1:k-1)}$ and the context of the $d^{th}$ document. The order of the keyphrases for the same document are randomly shuffled in each epoch of training.

We use Adam \cite{kingma2014adam} with a mini-batch size of $n = 256$ to optimize model parameters, with an initial learning rate $\alpha_1 = 0.002$ in the first 6,000 steps and $\alpha_2 = 0.0002$ in the rest steps. A gradient clipping $clip = 0.1$ is applied in each step. We also use early-stop in the training with a validation dataset. Model parameters are initialized by normal distributions \cite{glorot2010understanding}. Dropouts \cite{srivastava2014dropout} are applied on the word embedding with dropout rate $p = 0.1$ and on the GCN output with dropout rate $p = 0.5$ to reduce over-fitting. We also apply batch normalization \cite{ioffe2015batch} after the last graph convolutional layers to accelerate the training process.

In the decoding, we generate keyphrases based on the negative log generative probability of candidate keyphrases, with a beam-search in window size $= 100$  and search depth $= 5$. In practice, we find that the model tends to generate short keyphrases when using the negative log-likelihood as the keyphrase scores. Therefore, we propose a simple keyphrase length penalization, where we normalize the score of candidate keyphrase $\mathbf{y}$ by its length $|\mathbf{y}|$:
\begin{equation}
\bar{s}(\mathbf{y}) = \frac{s(\mathbf{y})}{\alpha + |\mathbf{y}|}
\end{equation}
where $s(\mathbf{y}) = -\log p(\mathbf{y})$ is the original score (negative log probability) and $\bar{s}$ is the normalized score. $\alpha$ is the length penalty factor, where larger $\alpha$ tends to generate shorter keyphrases, and smaller $\alpha$ generates longer keyphrases.
\begin{table}[t!]
\centering
\begin{tabular}{|c|c|c|c|} 
\hline
\textbf{Dataset} & \textbf{\#paper} & \textbf{\#keyphrase} & \textbf{length}\\
\hline
\multicolumn{4}{|c|}{\textbf{training data}}\\
\hline
Kp20k & 527830 & 2.94 & 1.80\\
\hline
\multicolumn{4}{|c|}{\textbf{validation data}}\\
\hline
Inspec & 1500 & 7.39 & 2.28\\
\hline
NUS &  \multicolumn{3}{|c|}{five-fold cross-validation}\\
\hline
SemEval& 188 & 3.84 & 1.97 \\
\hline
Krapivin& 1904 & 2.52 & 1.94\\
\hline
Kp20k& 20,000 & 2.94 & 1.80 \\
\hline
\multicolumn{4}{|c|}{\textbf{test data}}\\
\hline
Inspec & 500 & 7.70 & 2.28\\
\hline
NUS& 211 & 5.37 & 1.84\\
\hline
SemEval& 100 & 5.73 & 1.93\\
\hline
Krapivin& 400 & 3.24 & 2.01\\
\hline
Kp20k& 20,000 & 2.94 & 1.80\\
\hline
\end{tabular}
\caption{Statistics of five datasets. We use the original training data from the Inspec, SemEval, and Krapivin data sets as validation data to select the best length penalty factor $\alpha$.}
\label{tab:statistics}
\end{table}

\begin{table*}[t]
\centering
\setlength{\tabcolsep}{1.2mm}{
\begin{tabular}{|c|cc|cc|cc|cc|cc|} 
\hline
& \multicolumn{2}{|c|}{\textbf{Inspec}} & \multicolumn{2}{|c|}{\textbf{NUS}} & \multicolumn{2}{|c|}{\textbf{SemEval}} &  \multicolumn{2}{|c|}{\textbf{Krapivin}} &  \multicolumn{2}{|c|}{\textbf{Kp20k}}\\
&  $\mathbf{F_1}$\textbf{@5}& $\mathbf{F_1}$\textbf{@10} & $\mathbf{F_1}$\textbf{@5}& $\mathbf{F_1}$\textbf{@10} & $\mathbf{F_1}$\textbf{@5}& $\mathbf{F_1}$\textbf{@10} & $\mathbf{F_1}$\textbf{@5}& $\mathbf{F_1}$\textbf{@10} & $\mathbf{F_1}$\textbf{@5}& $\mathbf{F_1}$\textbf{@10} \\
\hline

\multicolumn{11}{|c|}{unsupervised methods}\\
\hline
Tf-Idf & 0.223 & 0.304 & 0.139 & 0.181 & 0.120 & 0.184 & 0.113 & 0.143 & 0.105 & 0.130\\
\hline
TextRank & 0.229 & 0.275 & 0.195 & 0.190 & 0.172 & 0.181 & 0.172 & 0.147 & 0.180 & 0.150\\
\hline
SingleRank & 0.214 & 0.297 & 0.145 & 0.169 & 0.132 & 0.169 & 0.096 & 0.137 & 0.099 & 0.124\\
\hline
ExpandRank & 0.211 & 0.295 & 0.137 & 0.162 & 0.135 & 0.163 & 0.096 & 0.136 & N/A & N/A\\
\hline
\hline
\multicolumn{11}{|c|}{supervised methods}\\
\hline
RNN & 0.000 & 0.000 & 0.005 & 0.004 & 0.004 & 0.003 & 0.002 & 0.001 & 0.138 & 0.009\\
\hline
CopyRNN & 0.292 & 0.336 & 0.342 & 0.317 & 0.291 & 0.296 & 0.302 & 0.252 & 0.328 & 0.255\\
\hline
CNN & 0.088 & 0.069 & 0.176 & 0.133 & 0.162 & 0.127 & 0.141 & 0.098 & 0.188 & 0.203\\
\hline
CopyCNN & 0.285 & 0.346 & 0.342 & 0.330 & 0.295 & 0.308 & 0.314 & 0.272 & 0.351 & 0.288\\
\hline
\hline
\rnnmethod{} & 0.347 & 0.386 & 0.383 & 0.345 & 0.328 & 0.324 & 0.318 & 0.274 & 0.333 & 0.281\\
\hline
\nodivmethod{} & 0.375*& 0.387& 0.421*& 0.375**& 0.377*& 0.350**& 0.340**& 0.280**& 0.341**& 0.282*\\
\hline
\method{} & \textbf{0.386}** & \textbf{0.417}** & \textbf{0.460}** & \textbf{0.402}** & \textbf{0.401}** & \textbf{0.389}** & \textbf{0.363}** & \textbf{0.297}** & \textbf{0.368}** & \textbf{0.292}**\\
\hline
\hline
\multicolumn{11}{|c|}{\textbf{Normalized Discounted Cumulative Gain @10 (NDCG@10)}}\\
\hline
\rnnmethod{} & \multicolumn{2}{|c|}{0.448} & \multicolumn{2}{|c|}{0.501} & \multicolumn{2}{|c|}{0.440} & \multicolumn{2}{|c|}{0.476} & \multicolumn{2}{|c|}{0.463}\\
\hline
\nodivmethod{} & \multicolumn{2}{|c|}{0.479} & \multicolumn{2}{|c|}{0.536} & \multicolumn{2}{|c|}{0.482} & \multicolumn{2}{|c|}{0.499} & \multicolumn{2}{|c|}{0.498}\\
\hline
\method{} & \multicolumn{2}{|c|}{\textbf{0.503}} & \multicolumn{2}{|c|}{\textbf{0.591}} & \multicolumn{2}{|c|}{\textbf{0.518}} & \multicolumn{2}{|c|}{\textbf{0.534}} & \multicolumn{2}{|c|}{\textbf{0.532}}\\
\hline
\end{tabular}}
\caption{The performance of keyphrase extraction on five benchmarks. The results of the former six methods are taken from \cite{meng2016knowledge} and the results of CNN and CopyCNN are taken from \cite{zhang2017deep}. The significance levels (** 0.01, * 0.1) between different methods (GraphPointer v.s SeqPointer, DivGraphPointer v.s GraphPointer) are also provided.}
\label{tab:performance}
\end{table*}

\section{Experiments}

\subsection{Experimental Setting}

\paragraph{Data}
We use the data set Kp20k \cite{meng2017deep} for training. Kp20k contains a large amount of high-quality scientific metadata in the computer science domain from various online digital libraries \cite{meng2016knowledge}. We follow the official setting of this dataset and split the dataset into training (527,830 articles), validation (20,000 articles) and test (20,000 articles) data. We further test the model trained with KP20k on four widely-adopted keyphrase extraction data sets including Inspec \cite{hulth2003improved}, NUS \cite{nguyen2007keyphrase}, SemEval-2010 \cite{kim2010semeval} and Krapivin \cite{krapivin2009large}. Following \cite{meng2017deep}, we take the concatenation of the title and the abstract as the content of a paper for all these datasets. No text pre-processing steps are conducted. In this paper, we focus on keyphrase extraction. Therefore, only the keyphrases that appear in the documents are used for training and evaluation. Table \ref{tab:statistics} provides the statistics on the number of papers, the average number of keyphrases and the corresponding average keyphrase length for each benchmark datasets.





\paragraph{Model Setting} 
A 3-layer Gated Recurrent Unit (GRU) \cite{chung2014empirical} is used as the RNN recurrent function. The dimension of both the input and the output of GRU is set to 400, while the word embedding dimension is set to 300. 
The number of GCN layers is empirically set to 6. The length penalty factor $\alpha$ is selected according to the validations on different data sets. All the other hyper-parameters are the same when we evaluate our different models on different datasets. The word embeddings (or the node embeddings) are fixed to the pre-trained fastText model \cite{bojanowski2016enriching}, which is trained on Wikipedia 2017, UMBC web-base corpus and statmt.org news dataset (16B tokens). It breaks words into sub-words, which can help handle the problem of out-of-vocabulary (OOV).

\paragraph{Baseline}
We compare our models with four supervised algorithms:  RNN \cite{meng2017deep}, CopyRNN \cite{meng2017deep}, CNN\cite{zhang2017deep}, and CopyCNN\cite{zhang2017deep}. Considering that the unsupervised ranking-based methods motivate our proposed graph-based encoder solution, we also compare our models with four well-known unsupervised algorithms for keyphrase extraction including Tf-Idf \cite{sparck1972statistical}, TextRank \cite{mihalcea2004textrank}, SingleRank \cite{wan2008single}, ExpandRank \cite{wan2008single}.

We also compare several different variants of our algorithms. We compare with the method of encoding documents with 2-layer bi-directional LSTM \cite{hochreiter1997long} and decoding the keyphrases with pointer networks, marked as \rnnmethod{}. We also compare with the method with graph-based encoder and vanilla pointer decoder, marked as \nodivmethod{}. No diversity mechanisms are used during decoding for both \rnnmethod{} and \nodivmethod{}. The hyper-parameters of these two models are also selected according to the validation data.



\paragraph{Evaluation} 
Following the literature, the macro-averaged precision, recall and F1 measures are used to measure the overall performance. Here, precision is defined as the number of correctly-predicted keyphrases over the number of all predicted keyphrases; recall is computed as the number of correctly predicted keyphrases over the total number of data records, and F1 is the harmonic average of precision and recall.

Besides the set-based metrics, we also evaluate our models with a rank-based metric, Normalized Discounted Cumulative Gain (NDCG) \cite{wang2013theoretical}. Since most keyphrase extraction models' output is sequential, we believe that the rank-based metric could better measure the performance of those models.

In the evaluation, we apply Porter Stemmer \cite{porter1980algorithm} to both target keyphrases and predicted keyphrases when determining the match of keyphrases and match of the identical word. The embedding of different variants of an identical word are averaged when fed as the input of graph-based encoder.


\subsection{Results}

The performance of different algorithms on five benchmarks are summarized in Table \ref{tab:performance}. For each method, the table presents the performance of generating 5 and 10 keyphrases with respect to F1 measure. We also include the truncated NDCG measure of generating 10 keyphrases in the table. The best results are highlighted in bold. We can see that for most cases (except RNN and CNN), the supervised models outperform all the unsupervised algorithms. This is not surprising since the supervised models are trained end-to-end with supervised data. The RNN model and CNN do not perform well on nearly all datasets (e.g., Inspec, NUS, SemEval, and Krapivin). The reason is that they cannot generate words that are not in the vocabulary (OOV words) during the decoding process, while in this paper we only allow to generate words appeared in the given documents. This problem is addressed by the \rnnmethod{} model, which utilizes the pointer network to copy words from the source text, and hence the performance is significantly improved. We can also see such phenomenon on CopyCNN.

Comparing SeqPointer and CopyRNN, we can observe that SeqPointer outperforms CopyRNN on all data sets. This could be due to the fact that generation mechanism interferes with the copy mechanism in CopyRNN. Implementation details could also contribute to the performance difference. For example, we utilize fastText as word embedding to handle the OOV problem, while \citet{meng2017deep} randomly initialize their word embeddings. Moreover, We use a length penalty mechanism to solve  the problem of short phrase generation, while \citet{meng2017deep} applied a simple heuristic by preserving only the first single-word phrase and removing the rest. We also use different hidden dimension, learning rate, dropout rate, beam depth, and beam size settings.

By replacing the RNN encoder with the graph convolutional network encoder, the performance of GraphPointer is further improved. This shows that although sequence-based encoders are effective at capturing sequential information such as word order and adjacency, such sequential information is not sufficient in the keyphrase extraction task. It turns out that the long-term dependency and information aggregation are more important for precisely extracting keyphrases.

Our proposed model \method{} achieves the best performance by modeling document-level word salience during the encoding process (the graph encoder), and increasing the diversity of keyphrases during decoding with diversified pointer networks. 




\begin{table}[t!]
\centering
\begin{tabular}{|c|cc|cc|} 
\hline
& \multicolumn{2}{|c|}{\textbf{Inspec}} & \multicolumn{2}{|c|}{\textbf{Krapivin}} \\
&  $\mathbf{F_1}$\textbf{@5}& $\mathbf{F_1}$\textbf{@10} & $\mathbf{F_1}$\textbf{@5}& $\mathbf{F_1}$\textbf{@10} \\
\hline
\rnnmethod{} & 0.347 & 0.386 &  0.318 & 0.274\\
\hline
\hline
1-layer GCN& 0.351 & 0.365 & 0.320 & 0.261\\
\hline
3-layer GCN& 0.373 & 0.382 & 0.334 & 0.268\\
\hline
6-layer GCN& \textbf{0.375} & 0.387 & 0.340 & 0.280 \\
\hline
9-layer GCN& 0.374 & \textbf{0.397} & \textbf{0.362} & \textbf{0.284} \\
\hline
12-layer GCN& 0.373 & 0.394 & 0.344 & 0.283 \\
\hline
\end{tabular}
\caption{Effectiveness of encoding documents as graphs with graph convolutional neural networks.}
\label{table:gcn}
\end{table}

\begin{table}[t!]
\centering
\begin{tabular}{|l|cc|cc|} 
\hline
& \multicolumn{2}{|c|}{\textbf{Inspec}} & \multicolumn{2}{|c|}{\textbf{Krapivin}} \\
&  $\mathbf{F_1}$\textbf{@5}& $\mathbf{F_1}$\textbf{@10} & $\mathbf{F_1}$\textbf{@5}& $\mathbf{F_1}$\textbf{@10} \\
\hline
+neither & 0.375 & 0.387 & 0.340 & 0.280 \\
\hline
+coverage & 0.363 & 0.381 & 0.356 & 0.283\\
\hline
+context & 0.379 & 0.400 & 0.360 & 0.290\\
\hline
+both & \textbf{0.386} & \textbf{0.417} & \textbf{0.363} & \textbf{0.297}\\
\hline
\end{tabular}
\caption{Investigating the performance of \method{} with  different diversity mechanisms. Here "coverage" denotes coverage attention, and "context" denotes context modification.}
\label{table:corr}
\end{table}

\begin{table}[t!]
\centering
\begin{tabular}{|l|cc|cc|} 
\hline
& \multicolumn{2}{|c|}{\textbf{AIC@5}} & \multicolumn{2}{|c|}{\textbf{AIC@10}} \\
\hline
Inspec  & 0.057&\textbf{0.034} &0.062&\textbf{0.048}\\
\hline
NUS  &0.056&\textbf{0.041}& 0.063&\textbf{0.054}\\
\hline
SemEval  &0.048&\textbf{0.031} &0.051&\textbf{0.044}\\
\hline
Krapvin  &0.047&\textbf{0.033}& 0.050&\textbf{0.044}\\
\hline
Kp20k  &0.049&\textbf{0.030} &0.053&\textbf{0.040}\\
\hline
\end{tabular}
\caption{The Average Index of Coincidence (AIC@N) of the extracted keyphrases from GraphPointer (left) and DivGraphPointer (right) on five benchmarks}
\label{table:aic}
\end{table}

\begin{table}[t!]
\centering
\begin{tabular}{|l|cc|cc|} 
\hline
& \multicolumn{2}{|c|}{\textbf{Inspec}} & \multicolumn{2}{|c|}{\textbf{Krapivin}} \\
&  $\mathbf{F_1}$\textbf{@5}& $\mathbf{F_1}$\textbf{@10} & $\mathbf{F_1}$\textbf{@5}& $\mathbf{F_1}$\textbf{@10} \\
\hline
$\alpha = 0$ & \textbf{0.394} & 0.411  & 0.348 & 0.282 \\
\hline
$\alpha = 1$ & 0.386 & \textbf{0.417}  & 0.363 & \textbf{0.297} \\
\hline
$\alpha = 2$ & 0.379 & 0.415 & 0.365 & \textbf{0.297} \\
\hline
$\alpha = 5$ & 0.364 & 0.400 & \textbf{0.370} & 0.296 \\
\hline
$\alpha = 20$ & 0.349 & 0.389 & 0.363 & 0.295 \\
\hline
$\alpha = 100$  &   0.341 & 0.382  & 0.362 & 0.296 \\
\hline
\end{tabular}
\caption{The performance of \method{} w.r.t. the length penalty factor $\alpha$.}
\label{table:length}
\end{table}

\begin{figure*}[t!]
\center
\includegraphics[width=0.98\textwidth]{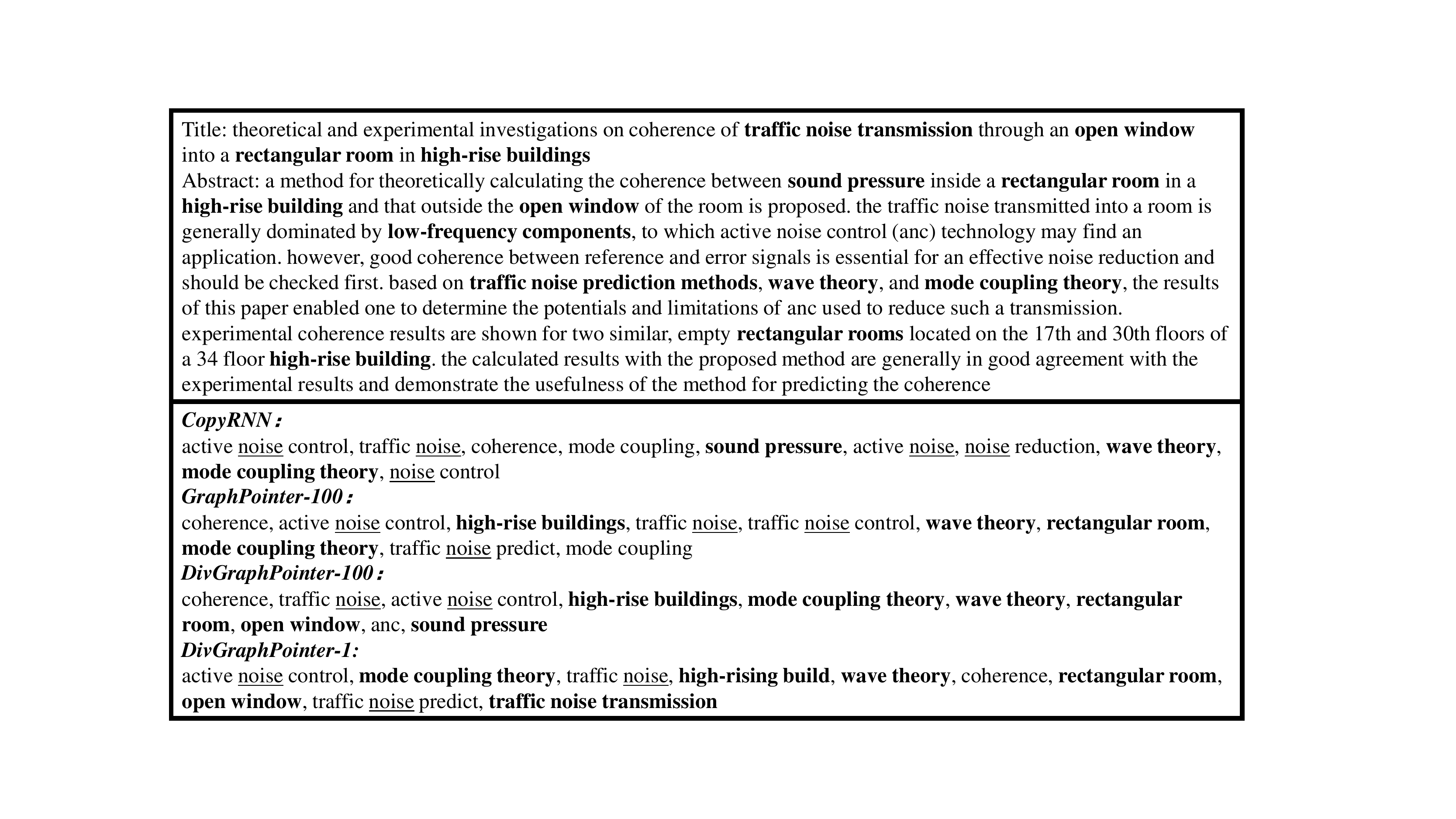}
\caption{An example of keyphrase extraction results with CopyRNN and our models. Phrases in bold are true keyphrases that predicted by the algorithms.}
\label{fig:example}
\end{figure*}

\subsection{Model Analysis}

We further conducted a detailed analysis of the proposed \method{} model.  We take two data sets Inspec and Krapivin as examples.


\subsubsection{Effectiveness of Encoding Documents as Graphs}
In this part, we evaluate the effectiveness of encoding documents with graphs. We compare  \rnnmethod{} and \nodivmethod{} with different numbers of graph convolutional layers. To focus on comparing the effectiveness of different encoders, we do not use the diversity mechanism and set the length penalty factor $\alpha = 1$  in all the compared algorithms.


The results on Inspec and Krapivin are summarized in Table \ref{table:gcn}. First, we can see that encoding documents as graphs significantly outperforms
the \rnnmethod{}, which encodes documents with bi-directional RNN, especially when more GCN layers are used. Especially, we notice that even 1-layer GCN can outperform RNN encoder on some metrics. This shows that the graph-based representation are very suitable for the keyphrase extraction task and demonstrates the effectiveness of graph-based encoder in aggregating information of multiple appearances of identical words and modeling long-term dependency.

Increasing the number of layers will increase the performance in the beginning. If too many layers are used, the performance will decrease due to over-fitting. In all our other experiments, we choose the number of graph convolutional layers as six by balancing the effectiveness and efficiency of the graph convolutional networks.



\subsubsection{Diversity Mechanism}

Next, we investigate the effectiveness of the proposed diversity mechanisms: context modification and coverage attention. We compare the following \method{} variants:
(1) with neither of them, (2) with only context modification, (3) with only coverage attention and (4) with both of them . Here we  also set the length penalty factor $\alpha = 1$. The results are presented in Table \ref{table:corr}.

We can see that the performance of adding context modification significantly improves comparing to the vanilla pointer networks. The results of adding coverage attention are mixed. On the Krapivin data set, the performance improves while on the Inspec data set, the performance decreases. The performance of adding both mechanism are very robust, significantly better than vanilla pointer networks. This shows that  the coverage attention and context modification focus on lexical-level and semantic-level duplication problem, respectively, thus are complementary to each other.

We also provide the Average Index of Coincidence (AIC) \cite{friedman1922index} of the extracted keyphrases from GraphPointer and DivGraphPointer in Table \ref{table:aic}. AIC represents the probability of the identical words appearing in the extracted keyphrases.  A smaller AIC indicates smaller redundancy. We can find that our proposed mechanism is very effective in improving keyphrase diversity. We also observe that this effect is more significant for top 5 keyphrases than top 10 keyphrases. It shows that the diversification mechanism tends to take effect at the begin of the keyphrase extraction process.



\subsubsection{Length Penalty Factor}
Fianlly, we investigate how the length penalty factor $\alpha$ affects the performance of the \method{}. Results with different values of length penalty factor $\alpha$ are presented in Table \ref{table:length}. Results show that the length penalty factors affect the model performance significantly. Either a small value of $\alpha$ (e.g., $\alpha$=0), which tends to generate long phrase, or a big value of $\alpha$ (e.g., $\alpha=100$), which tends to generate short phrases, yields inferior results. The best choice of $\alpha$ is between 0 and 1 for Inspec and between 1 and 5 for Krapivin, which indicates that both extreme cases - totally normalization $\alpha = 0$ or no normalization $\alpha = 100$ - are not good choices.


\begin{figure}[t!]
\center
\includegraphics[width=0.45\textwidth]{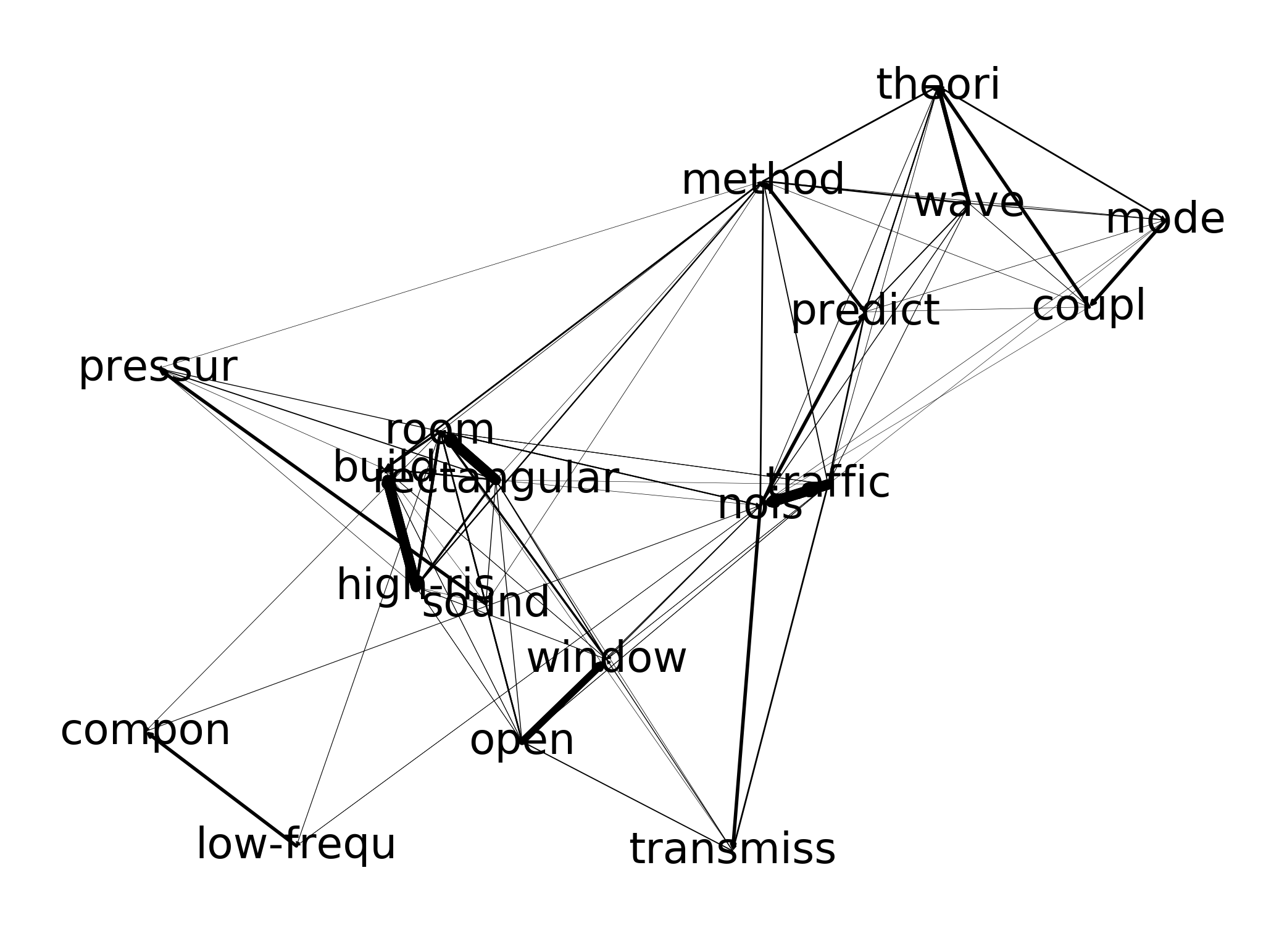}
\caption{A sub-graph on keyphrase words of the full word graph for the document in Figure \ref{fig:example}. The edge width is proportional to the edge weight.}
\label{fig:graph}
\end{figure}

\subsection{Case Analysis}
Finally, we present a case study on the results of the keyphrases extracted by different algorithms. Figure \ref{fig:example} presents the results of CopyRNN \cite{meng2017deep} and different variants of our algorithms.  "Model-$\alpha$" means that the length penalty factor $\alpha$ is used in the evaluated model. We also visualize a sub-graph on the words that appear in the keyphrases for clear illustration of encoding phase of graph-based methods. From the figures, we have the following observations:

\begin{enumerate}
\item The diversity mechanism is quite effective to increase the diversity of the generated keyphrases. For example, ``CopyRNN"  generates five similar keyphrases, which all contain the word ``noise", ``\nodivmethod{}-100" generates four such keyphrases, while ``\method{}-100" and ``\method{}-1" only generates two.

\item The length penalty factor $\alpha$ can efficiently control the granularity of generated keyphrases. For example, the keyphrase with only a single word ``coherence" ranks the first in 
``\method{}-100", but only ranks the sixth in ``\method{}-1", when the same trained model is used. 

\item The graph-based encoders can better estimate the salience of words by their relations. For example, ``high-rise building'' and ``rectangular room'' are highly relevant in the word graph, and thus are selected by the ``\nodivmethod{}'' model, while ``CopyRNN'' finds ``sound pressure'', which appears only once and only weakly connected to other keyphrases. The strong connection between ``traffic'' and ``noise'' also explains why the graph-based methods rank ``traffic noise'' higher than true keyphrase ``traffic noise transmission'' or ``traffic noise prediction methods''.
\end{enumerate}

\section{Conclusions and Future Work}
In this paper, we propose an end-to-end method called \method{} for extracting diverse keyphrases. It formulates documents as graphs and applies graph convolutional networks for encoding the graphs, which efficiently capture the document-level word salience by modeling both the short- and long-range dependency between words in documents and aggregate the information of multiple appearances of identical words.  To avoid extracting similar keyphrases, a diversified pointer network is proposed to generate diverse keyphrases from the nodes of the graphs. Experiments on five benchmark data sets show that our proposed \method{} model achieves state-of-the-art performance, significantly outperforming existing state-of-the-art supervised and unsupervised methods.


Our research can be extended in many directions. To begin with, currently our diversified pointer network decoders extract keyphrase in an auto-regressive fashion. We could  further leverage reinforcement learning to address the exposure bias as well as consequent error propagation problem in the sequential generation process. Moreover, our graph convolutional network encoder aggregates the word relation information through manually designed edge weights based on proximity at present. We would like to further explore utilizing graph attention networks (GATs) \cite{velickovic2017graph} to dynamically capture correlation between words in word graph. Finally, utilizing linguistic information when constructing edges in word graphs to ease keyphrase extraction is also an interesting future direction.



\section*{acknowledgements}
We would like to thank Rui Meng for sharing the source code and giving helpful advice.

%
\bibliographystyle{ACM-Reference-Format}
\bibliography{sample-base}


\begin{thebibliography}{48}


\ifx \showCODEN    \undefined \def \showCODEN     #1{\unskip}     \fi
\ifx \showDOI      \undefined \def \showDOI       #1{#1}\fi
\ifx \showISBNx    \undefined \def \showISBNx     #1{\unskip}     \fi
\ifx \showISBNxiii \undefined \def \showISBNxiii  #1{\unskip}     \fi
\ifx \showISSN     \undefined \def \showISSN      #1{\unskip}     \fi
\ifx \showLCCN     \undefined \def \showLCCN      #1{\unskip}     \fi
\ifx \shownote     \undefined \def \shownote      #1{#1}          \fi
\ifx \showarticletitle \undefined \def \showarticletitle #1{#1}   \fi
\ifx \showURL      \undefined \def \showURL       {\relax}        \fi
\providecommand\bibfield[2]{#2}
\providecommand\bibinfo[2]{#2}
\providecommand\natexlab[1]{#1}
\providecommand\showeprint[2][]{arXiv:#2}

\bibitem[\protect\citeauthoryear{Bahdanau, Cho, and Bengio}{Bahdanau
  et~al\mbox{.}}{2014}]%
        {bahdanau2014neural}
\bibfield{author}{\bibinfo{person}{Dzmitry Bahdanau},
  \bibinfo{person}{Kyunghyun Cho}, {and} \bibinfo{person}{Yoshua Bengio}.}
  \bibinfo{year}{2014}\natexlab{}.
\newblock \showarticletitle{Neural machine translation by jointly learning to
  align and translate}.
\newblock \bibinfo{journal}{\emph{arXiv preprint arXiv:1409.0473}}
  (\bibinfo{year}{2014}).
\newblock


\bibitem[\protect\citeauthoryear{Bastings, Titov, Aziz, Marcheggiani, and
  Sima'an}{Bastings et~al\mbox{.}}{2017}]%
        {bastings2017graph}
\bibfield{author}{\bibinfo{person}{Joost Bastings}, \bibinfo{person}{Ivan
  Titov}, \bibinfo{person}{Wilker Aziz}, \bibinfo{person}{Diego Marcheggiani},
  {and} \bibinfo{person}{Khalil Sima'an}.} \bibinfo{year}{2017}\natexlab{}.
\newblock \showarticletitle{Graph convolutional encoders for syntax-aware
  neural machine translation}.
\newblock \bibinfo{journal}{\emph{arXiv preprint arXiv:1704.04675}}
  (\bibinfo{year}{2017}).
\newblock


\bibitem[\protect\citeauthoryear{Bojanowski, Grave, Joulin, and
  Mikolov}{Bojanowski et~al\mbox{.}}{2016}]%
        {bojanowski2016enriching}
\bibfield{author}{\bibinfo{person}{Piotr Bojanowski}, \bibinfo{person}{Edouard
  Grave}, \bibinfo{person}{Armand Joulin}, {and} \bibinfo{person}{Tomas
  Mikolov}.} \bibinfo{year}{2016}\natexlab{}.
\newblock \showarticletitle{Enriching word vectors with subword information}.
\newblock \bibinfo{journal}{\emph{arXiv preprint arXiv:1607.04606}}
  (\bibinfo{year}{2016}).
\newblock


\bibitem[\protect\citeauthoryear{Bougouin, Boudin, and Daille}{Bougouin
  et~al\mbox{.}}{2013}]%
        {bougouin2013topicrank}
\bibfield{author}{\bibinfo{person}{Adrien Bougouin}, \bibinfo{person}{Florian
  Boudin}, {and} \bibinfo{person}{B{\'e}atrice Daille}.}
  \bibinfo{year}{2013}\natexlab{}.
\newblock \showarticletitle{Topicrank: Graph-based topic ranking for keyphrase
  extraction}. In \bibinfo{booktitle}{\emph{International Joint Conference on
  Natural Language Processing (IJCNLP)}}. \bibinfo{pages}{543--551}.
\newblock


\bibitem[\protect\citeauthoryear{Brin and Page}{Brin and Page}{1998}]%
        {brin1998anatomy}
\bibfield{author}{\bibinfo{person}{Sergey Brin} {and} \bibinfo{person}{Lawrence
  Page}.} \bibinfo{year}{1998}\natexlab{}.
\newblock \showarticletitle{The anatomy of a large-scale hypertextual web
  search engine}.
\newblock \bibinfo{journal}{\emph{Computer networks and ISDN systems}}
  \bibinfo{volume}{30}, \bibinfo{number}{1-7} (\bibinfo{year}{1998}),
  \bibinfo{pages}{107--117}.
\newblock


\bibitem[\protect\citeauthoryear{Carbonell and Goldstein}{Carbonell and
  Goldstein}{1998}]%
        {carbonell1998use}
\bibfield{author}{\bibinfo{person}{Jaime Carbonell} {and} \bibinfo{person}{Jade
  Goldstein}.} \bibinfo{year}{1998}\natexlab{}.
\newblock \showarticletitle{The use of MMR, diversity-based reranking for
  reordering documents and producing summaries}. In
  \bibinfo{booktitle}{\emph{Proceedings of the 21st annual international ACM
  SIGIR conference on Research and development in information retrieval}}. ACM,
  \bibinfo{pages}{335--336}.
\newblock


\bibitem[\protect\citeauthoryear{Chen, Zhang, Wu, Yan, and Li}{Chen
  et~al\mbox{.}}{2018}]%
        {chen2018keyphrase}
\bibfield{author}{\bibinfo{person}{Jun Chen}, \bibinfo{person}{Xiaoming Zhang},
  \bibinfo{person}{Yu Wu}, \bibinfo{person}{Zhao Yan}, {and}
  \bibinfo{person}{Zhoujun Li}.} \bibinfo{year}{2018}\natexlab{}.
\newblock \showarticletitle{Keyphrase generation with correlation constraints}.
\newblock \bibinfo{journal}{\emph{arXiv preprint arXiv:1808.07185}}
  (\bibinfo{year}{2018}).
\newblock


\bibitem[\protect\citeauthoryear{Chung, Gulcehre, Cho, and Bengio}{Chung
  et~al\mbox{.}}{2014}]%
        {chung2014empirical}
\bibfield{author}{\bibinfo{person}{Junyoung Chung}, \bibinfo{person}{Caglar
  Gulcehre}, \bibinfo{person}{KyungHyun Cho}, {and} \bibinfo{person}{Yoshua
  Bengio}.} \bibinfo{year}{2014}\natexlab{}.
\newblock \showarticletitle{Empirical evaluation of gated recurrent neural
  networks on sequence modeling}.
\newblock \bibinfo{journal}{\emph{arXiv preprint arXiv:1412.3555}}
  (\bibinfo{year}{2014}).
\newblock


\bibitem[\protect\citeauthoryear{Frank, Paynter, Witten, Gutwin, and
  Nevill-Manning}{Frank et~al\mbox{.}}{1999}]%
        {frank1999domain}
\bibfield{author}{\bibinfo{person}{Eibe Frank}, \bibinfo{person}{Gordon~W
  Paynter}, \bibinfo{person}{Ian~H Witten}, \bibinfo{person}{Carl Gutwin},
  {and} \bibinfo{person}{Craig~G Nevill-Manning}.}
  \bibinfo{year}{1999}\natexlab{}.
\newblock \showarticletitle{Domain-specific keyphrase extraction}. In
  \bibinfo{booktitle}{\emph{16th International joint conference on artificial
  intelligence (IJCAI 99)}}, Vol.~\bibinfo{volume}{2}. Morgan Kaufmann
  Publishers Inc., San Francisco, CA, USA, \bibinfo{pages}{668--673}.
\newblock


\bibitem[\protect\citeauthoryear{Friedman}{Friedman}{1922}]%
        {friedman1922index}
\bibfield{author}{\bibinfo{person}{William~Frederick Friedman}.}
  \bibinfo{year}{1922}\natexlab{}.
\newblock \bibinfo{booktitle}{\emph{The index of coincidence and its
  applications in cryptography}}.
\newblock \bibinfo{publisher}{Aegean Park Press}.
\newblock


\bibitem[\protect\citeauthoryear{Gehring, Auli, Grangier, Yarats, and
  Dauphin}{Gehring et~al\mbox{.}}{2017}]%
        {gehring2017convolutional}
\bibfield{author}{\bibinfo{person}{Jonas Gehring}, \bibinfo{person}{Michael
  Auli}, \bibinfo{person}{David Grangier}, \bibinfo{person}{Denis Yarats},
  {and} \bibinfo{person}{Yann~N Dauphin}.} \bibinfo{year}{2017}\natexlab{}.
\newblock \showarticletitle{Convolutional Sequence to Sequence Learning}.
\newblock \bibinfo{journal}{\emph{arXiv preprint arXiv:1705.03122}}
  (\bibinfo{year}{2017}).
\newblock


\bibitem[\protect\citeauthoryear{Glorot and Bengio}{Glorot and Bengio}{2010}]%
        {glorot2010understanding}
\bibfield{author}{\bibinfo{person}{Xavier Glorot} {and} \bibinfo{person}{Yoshua
  Bengio}.} \bibinfo{year}{2010}\natexlab{}.
\newblock \showarticletitle{Understanding the difficulty of training deep
  feedforward neural networks}. In \bibinfo{booktitle}{\emph{Proceedings of the
  Thirteenth International Conference on Artificial Intelligence and
  Statistics}}. \bibinfo{pages}{249--256}.
\newblock


\bibitem[\protect\citeauthoryear{Gu, Lu, Li, and Li}{Gu et~al\mbox{.}}{2016}]%
        {gu2016incorporating}
\bibfield{author}{\bibinfo{person}{Jiatao Gu}, \bibinfo{person}{Zhengdong Lu},
  \bibinfo{person}{Hang Li}, {and} \bibinfo{person}{Victor~OK Li}.}
  \bibinfo{year}{2016}\natexlab{}.
\newblock \showarticletitle{Incorporating copying mechanism in
  sequence-to-sequence learning}.
\newblock \bibinfo{journal}{\emph{arXiv preprint arXiv:1603.06393}}
  (\bibinfo{year}{2016}).
\newblock


\bibitem[\protect\citeauthoryear{Hasan and Ng}{Hasan and Ng}{2014}]%
        {hasan2014automatic}
\bibfield{author}{\bibinfo{person}{Kazi~Saidul Hasan} {and}
  \bibinfo{person}{Vincent Ng}.} \bibinfo{year}{2014}\natexlab{}.
\newblock \showarticletitle{Automatic keyphrase extraction: A survey of the
  state of the art}. In \bibinfo{booktitle}{\emph{Proceedings of the 52nd
  Annual Meeting of the Association for Computational Linguistics (Volume 1:
  Long Papers)}}, Vol.~\bibinfo{volume}{1}. \bibinfo{pages}{1262--1273}.
\newblock


\bibitem[\protect\citeauthoryear{He, Zhang, Ren, and Sun}{He
  et~al\mbox{.}}{2016}]%
        {he2016deep}
\bibfield{author}{\bibinfo{person}{Kaiming He}, \bibinfo{person}{Xiangyu
  Zhang}, \bibinfo{person}{Shaoqing Ren}, {and} \bibinfo{person}{Jian Sun}.}
  \bibinfo{year}{2016}\natexlab{}.
\newblock \showarticletitle{Deep residual learning for image recognition}. In
  \bibinfo{booktitle}{\emph{Proceedings of the IEEE conference on computer
  vision and pattern recognition}}. \bibinfo{pages}{770--778}.
\newblock


\bibitem[\protect\citeauthoryear{Hochreiter and Schmidhuber}{Hochreiter and
  Schmidhuber}{1997}]%
        {hochreiter1997long}
\bibfield{author}{\bibinfo{person}{Sepp Hochreiter} {and}
  \bibinfo{person}{J{\"u}rgen Schmidhuber}.} \bibinfo{year}{1997}\natexlab{}.
\newblock \showarticletitle{Long short-term memory}.
\newblock \bibinfo{journal}{\emph{Neural computation}} \bibinfo{volume}{9},
  \bibinfo{number}{8} (\bibinfo{year}{1997}), \bibinfo{pages}{1735--1780}.
\newblock


\bibitem[\protect\citeauthoryear{Hulth}{Hulth}{2003}]%
        {hulth2003improved}
\bibfield{author}{\bibinfo{person}{Anette Hulth}.}
  \bibinfo{year}{2003}\natexlab{}.
\newblock \showarticletitle{Improved automatic keyword extraction given more
  linguistic knowledge}. In \bibinfo{booktitle}{\emph{Proceedings of the 2003
  conference on Empirical methods in natural language processing}}. Association
  for Computational Linguistics, \bibinfo{pages}{216--223}.
\newblock


\bibitem[\protect\citeauthoryear{Ioffe and Szegedy}{Ioffe and Szegedy}{2015}]%
        {ioffe2015batch}
\bibfield{author}{\bibinfo{person}{Sergey Ioffe} {and}
  \bibinfo{person}{Christian Szegedy}.} \bibinfo{year}{2015}\natexlab{}.
\newblock \showarticletitle{Batch normalization: Accelerating deep network
  training by reducing internal covariate shift}. In
  \bibinfo{booktitle}{\emph{International Conference on Machine Learning}}.
  \bibinfo{pages}{448--456}.
\newblock


\bibitem[\protect\citeauthoryear{Kim, Medelyan, Kan, and Baldwin}{Kim
  et~al\mbox{.}}{2010}]%
        {kim2010semeval}
\bibfield{author}{\bibinfo{person}{Su~Nam Kim}, \bibinfo{person}{Olena
  Medelyan}, \bibinfo{person}{Min-Yen Kan}, {and} \bibinfo{person}{Timothy
  Baldwin}.} \bibinfo{year}{2010}\natexlab{}.
\newblock \showarticletitle{Semeval-2010 task 5: Automatic keyphrase extraction
  from scientific articles}. In \bibinfo{booktitle}{\emph{Proceedings of the
  5th International Workshop on Semantic Evaluation}}. Association for
  Computational Linguistics, \bibinfo{pages}{21--26}.
\newblock


\bibitem[\protect\citeauthoryear{Kim, Kim, Cattle, Otmakhova, Park, and
  Shin}{Kim et~al\mbox{.}}{2013}]%
        {kim2013applying}
\bibfield{author}{\bibinfo{person}{Youngsam Kim}, \bibinfo{person}{Munhyong
  Kim}, \bibinfo{person}{Andrew Cattle}, \bibinfo{person}{Julia Otmakhova},
  \bibinfo{person}{Suzi Park}, {and} \bibinfo{person}{Hyopil Shin}.}
  \bibinfo{year}{2013}\natexlab{}.
\newblock \showarticletitle{Applying graph-based keyword extraction to document
  retrieval}. In \bibinfo{booktitle}{\emph{Proceedings of the Sixth
  International Joint Conference on Natural Language Processing}}.
  \bibinfo{pages}{864--868}.
\newblock


\bibitem[\protect\citeauthoryear{Kingma and Ba}{Kingma and Ba}{2014}]%
        {kingma2014adam}
\bibfield{author}{\bibinfo{person}{Diederik Kingma} {and}
  \bibinfo{person}{Jimmy Ba}.} \bibinfo{year}{2014}\natexlab{}.
\newblock \showarticletitle{Adam: A method for stochastic optimization}.
\newblock \bibinfo{journal}{\emph{arXiv preprint arXiv:1412.6980}}
  (\bibinfo{year}{2014}).
\newblock


\bibitem[\protect\citeauthoryear{Kipf and Welling}{Kipf and Welling}{2016}]%
        {kipf2016semi}
\bibfield{author}{\bibinfo{person}{Thomas~N Kipf} {and} \bibinfo{person}{Max
  Welling}.} \bibinfo{year}{2016}\natexlab{}.
\newblock \showarticletitle{Semi-supervised classification with graph
  convolutional networks}.
\newblock \bibinfo{journal}{\emph{arXiv preprint arXiv:1609.02907}}
  (\bibinfo{year}{2016}).
\newblock


\bibitem[\protect\citeauthoryear{Krapivin, Autaeu, and Marchese}{Krapivin
  et~al\mbox{.}}{2009}]%
        {krapivin2009large}
\bibfield{author}{\bibinfo{person}{Mikalai Krapivin},
  \bibinfo{person}{Aliaksandr Autaeu}, {and} \bibinfo{person}{Maurizio
  Marchese}.} \bibinfo{year}{2009}\natexlab{}.
\newblock \bibinfo{booktitle}{\emph{Large dataset for keyphrases extraction}}.
\newblock \bibinfo{type}{{T}echnical {R}eport}.
  \bibinfo{institution}{University of Trento}.
\newblock


\bibitem[\protect\citeauthoryear{Li, Li, Ming, Hong, Tang, and Chua}{Li
  et~al\mbox{.}}{2010}]%
        {li2010question}
\bibfield{author}{\bibinfo{person}{Guangda Li}, \bibinfo{person}{Haojie Li},
  \bibinfo{person}{Zhaoyan Ming}, \bibinfo{person}{Richang Hong},
  \bibinfo{person}{Sheng Tang}, {and} \bibinfo{person}{Tat-Seng Chua}.}
  \bibinfo{year}{2010}\natexlab{}.
\newblock \showarticletitle{Question answering over community-contributed web
  videos}.
\newblock \bibinfo{journal}{\emph{IEEE MultiMedia}} \bibinfo{volume}{17},
  \bibinfo{number}{4} (\bibinfo{year}{2010}), \bibinfo{pages}{46--57}.
\newblock


\bibitem[\protect\citeauthoryear{Liu, Huang, Zheng, and Sun}{Liu
  et~al\mbox{.}}{2010}]%
        {liu2010automatic}
\bibfield{author}{\bibinfo{person}{Zhiyuan Liu}, \bibinfo{person}{Wenyi Huang},
  \bibinfo{person}{Yabin Zheng}, {and} \bibinfo{person}{Maosong Sun}.}
  \bibinfo{year}{2010}\natexlab{}.
\newblock \showarticletitle{Automatic keyphrase extraction via topic
  decomposition}. In \bibinfo{booktitle}{\emph{Proceedings of the 2010
  conference on empirical methods in natural language processing}}. Association
  for Computational Linguistics, \bibinfo{pages}{366--376}.
\newblock


\bibitem[\protect\citeauthoryear{Marcheggiani and Titov}{Marcheggiani and
  Titov}{2017}]%
        {marcheggiani2017encoding}
\bibfield{author}{\bibinfo{person}{Diego Marcheggiani} {and}
  \bibinfo{person}{Ivan Titov}.} \bibinfo{year}{2017}\natexlab{}.
\newblock \showarticletitle{Encoding sentences with graph convolutional
  networks for semantic role labeling}.
\newblock \bibinfo{journal}{\emph{arXiv preprint arXiv:1703.04826}}
  (\bibinfo{year}{2017}).
\newblock


\bibitem[\protect\citeauthoryear{Mei, Guo, and Radev}{Mei
  et~al\mbox{.}}{2010}]%
        {mei2010divrank}
\bibfield{author}{\bibinfo{person}{Qiaozhu Mei}, \bibinfo{person}{Jian Guo},
  {and} \bibinfo{person}{Dragomir Radev}.} \bibinfo{year}{2010}\natexlab{}.
\newblock \showarticletitle{Divrank: the interplay of prestige and diversity in
  information networks}. In \bibinfo{booktitle}{\emph{Proceedings of the 16th
  ACM SIGKDD international conference on Knowledge discovery and data mining}}.
  Acm, \bibinfo{pages}{1009--1018}.
\newblock


\bibitem[\protect\citeauthoryear{Meng, Han, Huang, He, and Brusilovsky}{Meng
  et~al\mbox{.}}{2016}]%
        {meng2016knowledge}
\bibfield{author}{\bibinfo{person}{Rui Meng}, \bibinfo{person}{Shuguang Han},
  \bibinfo{person}{Yun Huang}, \bibinfo{person}{Daqing He}, {and}
  \bibinfo{person}{Peter Brusilovsky}.} \bibinfo{year}{2016}\natexlab{}.
\newblock \showarticletitle{Knowledge-based content linking for online
  textbooks}. In \bibinfo{booktitle}{\emph{Web Intelligence (WI), 2016
  IEEE/WIC/ACM International Conference on}}. IEEE, \bibinfo{pages}{18--25}.
\newblock


\bibitem[\protect\citeauthoryear{Meng, Zhao, Han, He, Brusilovsky, and
  Chi}{Meng et~al\mbox{.}}{2017}]%
        {meng2017deep}
\bibfield{author}{\bibinfo{person}{Rui Meng}, \bibinfo{person}{Sanqiang Zhao},
  \bibinfo{person}{Shuguang Han}, \bibinfo{person}{Daqing He},
  \bibinfo{person}{Peter Brusilovsky}, {and} \bibinfo{person}{Yu Chi}.}
  \bibinfo{year}{2017}\natexlab{}.
\newblock \showarticletitle{Deep keyphrase generation}.
\newblock \bibinfo{journal}{\emph{arXiv preprint arXiv:1704.06879}}
  (\bibinfo{year}{2017}).
\newblock


\bibitem[\protect\citeauthoryear{Mihalcea and Tarau}{Mihalcea and
  Tarau}{2004}]%
        {mihalcea2004textrank}
\bibfield{author}{\bibinfo{person}{Rada Mihalcea} {and} \bibinfo{person}{Paul
  Tarau}.} \bibinfo{year}{2004}\natexlab{}.
\newblock \showarticletitle{Textrank: Bringing order into text}. In
  \bibinfo{booktitle}{\emph{Proceedings of the 2004 conference on empirical
  methods in natural language processing}}.
\newblock


\bibitem[\protect\citeauthoryear{Mikolov, Karafi{\'a}t, Burget,
  {\v{C}}ernock{\`y}, and Khudanpur}{Mikolov et~al\mbox{.}}{2010}]%
        {mikolov2010recurrent}
\bibfield{author}{\bibinfo{person}{Tom{\'a}{\v{s}} Mikolov},
  \bibinfo{person}{Martin Karafi{\'a}t}, \bibinfo{person}{Luk{\'a}{\v{s}}
  Burget}, \bibinfo{person}{Jan {\v{C}}ernock{\`y}}, {and}
  \bibinfo{person}{Sanjeev Khudanpur}.} \bibinfo{year}{2010}\natexlab{}.
\newblock \showarticletitle{Recurrent neural network based language model}. In
  \bibinfo{booktitle}{\emph{Eleventh Annual Conference of the International
  Speech Communication Association}}.
\newblock


\bibitem[\protect\citeauthoryear{Nguyen and Kan}{Nguyen and Kan}{2007}]%
        {nguyen2007keyphrase}
\bibfield{author}{\bibinfo{person}{Thuy~Dung Nguyen} {and}
  \bibinfo{person}{Min-Yen Kan}.} \bibinfo{year}{2007}\natexlab{}.
\newblock \showarticletitle{Keyphrase extraction in scientific publications}.
  In \bibinfo{booktitle}{\emph{International Conference on Asian Digital
  Libraries}}. Springer, \bibinfo{pages}{317--326}.
\newblock


\bibitem[\protect\citeauthoryear{Porter}{Porter}{1980}]%
        {porter1980algorithm}
\bibfield{author}{\bibinfo{person}{Martin~F Porter}.}
  \bibinfo{year}{1980}\natexlab{}.
\newblock \showarticletitle{An algorithm for suffix stripping}.
\newblock \bibinfo{journal}{\emph{Program}} \bibinfo{volume}{14},
  \bibinfo{number}{3} (\bibinfo{year}{1980}), \bibinfo{pages}{130--137}.
\newblock


\bibitem[\protect\citeauthoryear{Qazvinian, Radev, and {\"O}zg{\"u}r}{Qazvinian
  et~al\mbox{.}}{2010}]%
        {qazvinian2010citation}
\bibfield{author}{\bibinfo{person}{Vahed Qazvinian},
  \bibinfo{person}{Dragomir~R Radev}, {and} \bibinfo{person}{Arzucan
  {\"O}zg{\"u}r}.} \bibinfo{year}{2010}\natexlab{}.
\newblock \showarticletitle{Citation summarization through keyphrase
  extraction}. In \bibinfo{booktitle}{\emph{Proceedings of the 23rd
  international conference on computational linguistics}}. Association for
  Computational Linguistics, \bibinfo{pages}{895--903}.
\newblock


\bibitem[\protect\citeauthoryear{Santos, Macdonald, and Ounis}{Santos
  et~al\mbox{.}}{2010}]%
        {santos2010exploiting}
\bibfield{author}{\bibinfo{person}{Rodrygo~LT Santos}, \bibinfo{person}{Craig
  Macdonald}, {and} \bibinfo{person}{Iadh Ounis}.}
  \bibinfo{year}{2010}\natexlab{}.
\newblock \showarticletitle{Exploiting query reformulations for web search
  result diversification}. In \bibinfo{booktitle}{\emph{Proceedings of the 19th
  international conference on World wide web}}. ACM, \bibinfo{pages}{881--890}.
\newblock


\bibitem[\protect\citeauthoryear{See, Liu, and Manning}{See
  et~al\mbox{.}}{2017}]%
        {see2017get}
\bibfield{author}{\bibinfo{person}{Abigail See}, \bibinfo{person}{Peter~J Liu},
  {and} \bibinfo{person}{Christopher~D Manning}.}
  \bibinfo{year}{2017}\natexlab{}.
\newblock \showarticletitle{Get to the point: Summarization with
  pointer-generator networks}.
\newblock \bibinfo{journal}{\emph{arXiv preprint arXiv:1704.04368}}
  (\bibinfo{year}{2017}).
\newblock


\bibitem[\protect\citeauthoryear{Sparck~Jones}{Sparck~Jones}{1972}]%
        {sparck1972statistical}
\bibfield{author}{\bibinfo{person}{Karen Sparck~Jones}.}
  \bibinfo{year}{1972}\natexlab{}.
\newblock \showarticletitle{A statistical interpretation of term specificity
  and its application in retrieval}.
\newblock \bibinfo{journal}{\emph{Journal of documentation}}
  \bibinfo{volume}{28}, \bibinfo{number}{1} (\bibinfo{year}{1972}),
  \bibinfo{pages}{11--21}.
\newblock


\bibitem[\protect\citeauthoryear{Srivastava, Hinton, Krizhevsky, Sutskever, and
  Salakhutdinov}{Srivastava et~al\mbox{.}}{2014}]%
        {srivastava2014dropout}
\bibfield{author}{\bibinfo{person}{Nitish Srivastava},
  \bibinfo{person}{Geoffrey~E Hinton}, \bibinfo{person}{Alex Krizhevsky},
  \bibinfo{person}{Ilya Sutskever}, {and} \bibinfo{person}{Ruslan
  Salakhutdinov}.} \bibinfo{year}{2014}\natexlab{}.
\newblock \showarticletitle{Dropout: a simple way to prevent neural networks
  from overfitting.}
\newblock \bibinfo{journal}{\emph{Journal of machine learning research}}
  \bibinfo{volume}{15}, \bibinfo{number}{1} (\bibinfo{year}{2014}),
  \bibinfo{pages}{1929--1958}.
\newblock


\bibitem[\protect\citeauthoryear{Sutskever, Vinyals, and Le}{Sutskever
  et~al\mbox{.}}{2014}]%
        {sutskever2014sequence}
\bibfield{author}{\bibinfo{person}{Ilya Sutskever}, \bibinfo{person}{Oriol
  Vinyals}, {and} \bibinfo{person}{Quoc~V Le}.}
  \bibinfo{year}{2014}\natexlab{}.
\newblock \showarticletitle{Sequence to sequence learning with neural
  networks}. In \bibinfo{booktitle}{\emph{Advances in neural information
  processing systems}}. \bibinfo{pages}{3104--3112}.
\newblock


\bibitem[\protect\citeauthoryear{Tu, Lu, Liu, Liu, and Li}{Tu
  et~al\mbox{.}}{2016}]%
        {tu2016modeling}
\bibfield{author}{\bibinfo{person}{Zhaopeng Tu}, \bibinfo{person}{Zhengdong
  Lu}, \bibinfo{person}{Yang Liu}, \bibinfo{person}{Xiaohua Liu}, {and}
  \bibinfo{person}{Hang Li}.} \bibinfo{year}{2016}\natexlab{}.
\newblock \showarticletitle{Modeling coverage for neural machine translation}.
\newblock \bibinfo{journal}{\emph{arXiv preprint arXiv:1601.04811}}
  (\bibinfo{year}{2016}).
\newblock


\bibitem[\protect\citeauthoryear{Velickovic, Cucurull, Casanova, Romero, Lio,
  and Bengio}{Velickovic et~al\mbox{.}}{2017}]%
        {velickovic2017graph}
\bibfield{author}{\bibinfo{person}{Petar Velickovic}, \bibinfo{person}{Guillem
  Cucurull}, \bibinfo{person}{Arantxa Casanova}, \bibinfo{person}{Adriana
  Romero}, \bibinfo{person}{Pietro Lio}, {and} \bibinfo{person}{Yoshua
  Bengio}.} \bibinfo{year}{2017}\natexlab{}.
\newblock \showarticletitle{Graph attention networks}.
\newblock \bibinfo{journal}{\emph{arXiv preprint arXiv:1710.10903}}
  \bibinfo{volume}{1}, \bibinfo{number}{2} (\bibinfo{year}{2017}).
\newblock


\bibitem[\protect\citeauthoryear{Vinyals, Fortunato, and Jaitly}{Vinyals
  et~al\mbox{.}}{2015}]%
        {vinyals2015pointer}
\bibfield{author}{\bibinfo{person}{Oriol Vinyals}, \bibinfo{person}{Meire
  Fortunato}, {and} \bibinfo{person}{Navdeep Jaitly}.}
  \bibinfo{year}{2015}\natexlab{}.
\newblock \showarticletitle{Pointer networks}. In
  \bibinfo{booktitle}{\emph{Advances in Neural Information Processing
  Systems}}. \bibinfo{pages}{2692--2700}.
\newblock


\bibitem[\protect\citeauthoryear{Wan and Xiao}{Wan and Xiao}{2008}]%
        {wan2008single}
\bibfield{author}{\bibinfo{person}{Xiaojun Wan} {and} \bibinfo{person}{Jianguo
  Xiao}.} \bibinfo{year}{2008}\natexlab{}.
\newblock \showarticletitle{Single Document Keyphrase Extraction Using
  Neighborhood Knowledge.}. In \bibinfo{booktitle}{\emph{AAAI}},
  Vol.~\bibinfo{volume}{8}. \bibinfo{pages}{855--860}.
\newblock


\bibitem[\protect\citeauthoryear{Wang, Wang, Li, He, and Liu}{Wang
  et~al\mbox{.}}{2013}]%
        {wang2013theoretical}
\bibfield{author}{\bibinfo{person}{Yining Wang}, \bibinfo{person}{Liwei Wang},
  \bibinfo{person}{Yuanzhi Li}, \bibinfo{person}{Di He}, {and}
  \bibinfo{person}{Tie-Yan Liu}.} \bibinfo{year}{2013}\natexlab{}.
\newblock \showarticletitle{A theoretical analysis of NDCG type ranking
  measures}. In \bibinfo{booktitle}{\emph{Conference on Learning Theory}}.
  \bibinfo{pages}{25--54}.
\newblock


\bibitem[\protect\citeauthoryear{Yasunaga, Zhang, Meelu, Pareek, Srinivasan,
  and Radev}{Yasunaga et~al\mbox{.}}{2017}]%
        {yasunaga2017graph}
\bibfield{author}{\bibinfo{person}{Michihiro Yasunaga}, \bibinfo{person}{Rui
  Zhang}, \bibinfo{person}{Kshitijh Meelu}, \bibinfo{person}{Ayush Pareek},
  \bibinfo{person}{Krishnan Srinivasan}, {and} \bibinfo{person}{Dragomir
  Radev}.} \bibinfo{year}{2017}\natexlab{}.
\newblock \showarticletitle{Graph-based Neural Multi-Document Summarization}.
\newblock \bibinfo{journal}{\emph{arXiv preprint arXiv:1706.06681}}
  (\bibinfo{year}{2017}).
\newblock


\bibitem[\protect\citeauthoryear{Zhang, Wang, Gong, and Huang}{Zhang
  et~al\mbox{.}}{2016}]%
        {zhang2016keyphrase}
\bibfield{author}{\bibinfo{person}{Qi Zhang}, \bibinfo{person}{Yang Wang},
  \bibinfo{person}{Yeyun Gong}, {and} \bibinfo{person}{Xuanjing Huang}.}
  \bibinfo{year}{2016}\natexlab{}.
\newblock \showarticletitle{Keyphrase extraction using deep recurrent neural
  networks on Twitter}. In \bibinfo{booktitle}{\emph{Proceedings of the 2016
  Conference on Empirical Methods in Natural Language Processing}}.
  \bibinfo{pages}{836--845}.
\newblock


\bibitem[\protect\citeauthoryear{Zhang, Fang, and Weidong}{Zhang
  et~al\mbox{.}}{2017}]%
        {zhang2017deep}
\bibfield{author}{\bibinfo{person}{Yong Zhang}, \bibinfo{person}{Yang Fang},
  {and} \bibinfo{person}{Xiao Weidong}.} \bibinfo{year}{2017}\natexlab{}.
\newblock \showarticletitle{Deep keyphrase generation with a convolutional
  sequence to sequence model}. In \bibinfo{booktitle}{\emph{Systems and
  Informatics (ICSAI), 2017 4th International Conference on}}. IEEE,
  \bibinfo{pages}{1477--1485}.
\newblock


\bibitem[\protect\citeauthoryear{Zhang and Xiao}{Zhang and Xiao}{2018}]%
        {zhang2018keyphrase}
\bibfield{author}{\bibinfo{person}{Yong Zhang} {and} \bibinfo{person}{Weidong
  Xiao}.} \bibinfo{year}{2018}\natexlab{}.
\newblock \showarticletitle{Keyphrase Generation Based on Deep Seq2seq Model}.
\newblock \bibinfo{journal}{\emph{IEEE Access}}  \bibinfo{volume}{6}
  (\bibinfo{year}{2018}), \bibinfo{pages}{46047--46057}.
\newblock


\end{thebibliography}

\end{document}